\documentclass{svproc}

\usepackage{url}
\usepackage{graphicx}
\usepackage{amsmath}
\usepackage{tabularx}
\usepackage{multirow}
\usepackage{subfig}
\usepackage{booktabs}

\begin{document}
\pagestyle{empty}

\mainmatter              
\title{TERRA-CD: Multi-Temporal Framework for Multi-class and Semantic Change Detection}
\titlerunning{TERRA-CD}  
%
\author{Omkar Oak\inst{1} \and Rukmini Nazre\inst{2} \and Rujuta Budke\inst{3} \and Suraj Sawant\inst{4}}
\authorrunning{Omkar Oak et al.} 

%

\institute{COEP Technological University, Pune, India, \email{oakos21.comp@coeptech.ac.in}\\ \and 
University of Massachusetts, Amherst, USA, \email{rnazre@umass.edu}\\ \and
North Carolina State University, USA, \email{rabudke@ncsu.edu}\\ \and
COEP Technological University, Pune, India, \email{sts.comp@coeptech.ac.in}}

\maketitle

\vspace{-10pt}
\begin{abstract}
Urban vegetation monitoring plays a vital role in understanding environmental changes, yet comprehensive datasets for this purpose remain limited. To address this gap, we present the \textbf{Te}mporal \textbf{R}emote-sensing \textbf{R}epository for \textbf{A}nalyzing \textbf{C}hange \textbf{D}etection (TERRA-CD), a benchmark dataset comprising 5,221 Sentinel-2 image pairs from 2019 and 2024, covering 232 cities across the USA and Europe. The dataset features three distinct annotation schemes: 4-class land cover mapping masks, 3-class vegetation change masks, and 13-class semantic change masks capturing all possible land cover transitions. Using various deep learning approaches including Siamese networks, STANet variants, Bi-SRNet, Changemask, Post-Classification Comparison, and HRSCD strategies, we evaluated the dataset's effectiveness for both vegetation Multi-class Change Detection as well as Semantic Change Detection. The proposed dataset and methods are available at \url{https://github.com/omkarsoak/TERRA-CD}.

\keywords{Deep Learning, Image Processing, Remote Sensing, Semantic Change Detection, Semantic Segmentation}
\end{abstract}

\section{Introduction}
Urban vegetation change detection using satellite imagery plays a critical role in environmental monitoring and urban planning. While change detection has been extensively studied, most existing datasets focus on binary change detection \cite{chen2020spatial} or disaster damage assessment \cite{adriano2021learning}. The limited availability of multi-class change detection datasets, with notable exceptions being HRSCD \cite{daudt2019multitask} and SECOND \cite{yang2020semantic}, restricts the development of advanced change detection methodologies.\par
We present the \textbf{Te}mporal \textbf{R}emote-sensing \textbf{R}epository for \textbf{A}nalyzing \textbf{C}hange \textbf{D}etection (TERRA-CD), a large-scale Multi-class Change Detection (MCD) and Semantic Change Detection (SCD) dataset focused on urban vegetation and building patterns. The dataset consists of Sentinel-2 imagery spanning 232 cities across the USA and Europe with 10m resolution from 2019 and 2024. TERRA-CD supports two main purposes: Semantic Segmentation and Change Detection. It provides semantic segmentation masks for 4 classes, 3-class MCD masks, and 13-class SCD masks capturing all transitions between land cover classes. We benchmark TERRA-CD using traditional MCD and SCD architectures.

\section{Related Works}
Change detection tasks vary in complexity from binary classification identifying changed versus unchanged areas, to multi-class detection recognizing types of changes, to semantic analysis providing detailed ``from-to" transitions \cite{zhu2024review}.
\vspace{-20pt}
\begin{table}[h!]
\scriptsize
\setlength{\tabcolsep}{4.5pt}
\renewcommand{\arraystretch}{1.5}
\caption{Comparison of Available SCD Datasets}
\label{tab:dataset-comparison}
\begin{tabular}{@{\extracolsep{\fill}} lccccc}
\hline
\textbf{Dataset} & \textbf{Years} & \textbf{Resolution} & \textbf{Img. Pairs} &  \textbf{Application} & \textbf{Region} \\
\hline
Taizhou & 2000/2003 & 30m & 1 &   City Expansion & Taizhou \\
Nanjing & 2000/2002 & 30m & 1 &   City Expansion & Nanjing \\
Mts-WH & 2002/2009 & 1m & 1 & Urban Scene  & Wuhan, China \\
Yancheng & 2006/2007 & 30m & 2 & Vegetation  & Jiangsu, China \\
USA & 2004/2007 & 30m & 1 & Farmland  & Oregon, USA \\
HRSCD & 2005/2012 & 0.5m & 291 & Land Cover  & France \\
xBD & - & 0.3m & 11,034 & Building Damage & 15 countries \\
SECOND & - & 0.5-3m & 4,662 & Land Cover  & China \\
Hi-UCD & 2018/2019 & 0.1m & 40,800 & Land Cover  & Tallinn, Estonia \\
BDD & - & 0.5-1.5m & 1,147 & Building Damage & - \\
\hline
\textbf{TERRA-CD} & \textbf{2019/2024} & \textbf{10m} & \textbf{5,221} & \textbf{Vegetation} & \textbf{USA \& Europe} \\
\hline
\end{tabular}
\end{table}

As shown in Table \ref{tab:dataset-comparison}, existing SCD datasets have significant limitations. Early datasets like TAIZHOU \cite{lyu2016learning}, NANJING \cite{du2019unsupervised}, Mts-WH \cite{wu2016scene}, YANCHENG \cite{song2018change}, and USA \cite{liu2019review} contain extremely limited image pairs, sometimes just one, with restricted scope and resolution. More recent datasets including HRSCD \cite{daudt2019multitask}, SECOND \cite{yang2020semantic}, and Hi-UCD \cite{tian2022large} offer improvements but remain geographically concentrated, primarily in China, with limited coverage of the USA and Europe. Existing datasets focus exclusively on urban expansion or specific damage assessment, neglecting vegetation monitoring.

Change detection in remote sensing has evolved from traditional methods, using handcrafted features, to advanced deep learning approaches \cite{hussain2013change}. Modern architectures fall into three main categories: single-branch networks processing concatenated images through a unified pipeline, dual-branch systems employing parallel processing with feature fusion like Siamese networks, and multi-task structures combining change detection with classification tasks.  For our benchmarking, we evaluate MCD using Siamese networks, Post-Classification Comparison (PCC) methods, and STANet variants, while SCD is assessed using the four HRSCD strategies, Bi-SRNet and Changemask.

\section{TERRA-CD}
In this section, we present TERRA-CD, a benchmark dataset for detecting changes in urban vegetation. We detail the dataset's statistical composition, acquisition methodology, annotation framework, and strategic approaches for change mask generation.

\subsection{Dataset Statistics}
\begin{figure}[h!]
\centering
\includegraphics[width=\linewidth]{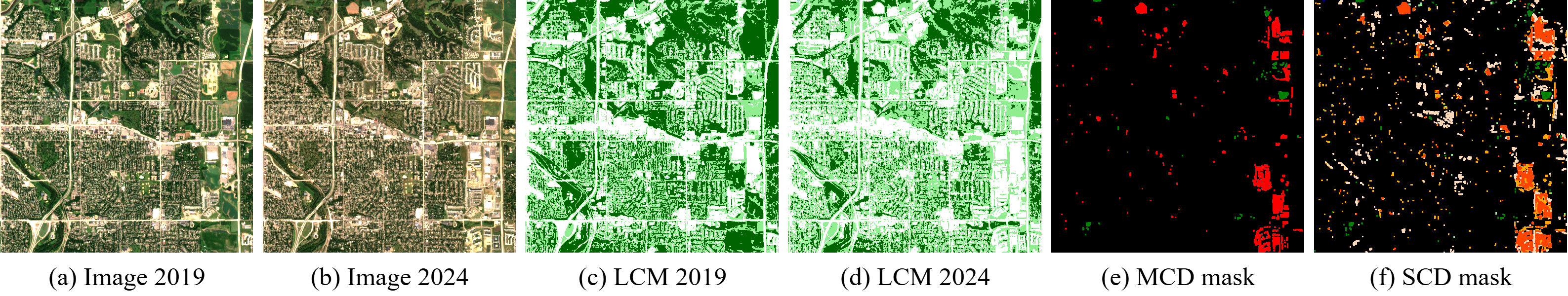}
\caption{TERRA-CD Dataset contents for a every tile}
\label{fig:dataset-sample}
\end{figure}

The TERRA-CD dataset comprises of 5,221 bitemporal image pairs of $256\times256$ pixels, with each pair consisting of one image from 2019 and one image from 2024. Figure \ref{fig:dataset-sample} illustrates a sample tile from the dataset. The images are captured at a resolution of 10 meters per pixel using Sentinel-2 imagery and downloaded in GeoTIFF format using Google Earth Engine's Python API. TERRA-CD includes images from 171 cities in the USA and 61 cities in Europe, with each city divided into nine spatial tiles to ensure sufficient land coverage.

\vspace{-10pt}
\vspace{-10pt}

The images are acquired during the months of July to September in 2019 and 2024, with a cloud cover threshold of 3\%. Each image pair is annotated with LCM masks containing four classes -- \textit{water, built area, sparse vegetation}, and \textit{dense vegetation}, MCD masks containing three classes -- \textit{no change, vegetation increase,} and \textit{vegetation decrease}, and SCD masks containing 13 classes covering all transitions as detailed in Figure \ref{tab:SCD-class-description}.

\vspace{-20pt}
\begin{table}[h!]
\small
\setlength\tabcolsep{3pt}
\centering
\caption{Change Detection class labels and pixel distrbutions. Row names and Column names represent LCM classes in 2019 and 2024 respectively.}
\label{tab:SCD-class-description}
\begin{tabular}{@{\extracolsep{\fill}} l c c c c}
\hline
2019 $\downarrow$ $\cdot$ 2024 $\rightarrow$ & Water & Built Area & Sparse Veg. & Dense Veg. \\
\hline
Water & 0 (-)     & 1 (0.17\%)  & 2 (0.02\%)   & 3 (0.01\%)  \\
Built Area & 4 (0.1\%)   & 0 (-)  & 5 (0.97\%)   & 6 (0.45\%)  \\
Sparse Veg. & 7 (0.01\%)  & 8 (1.61\%)  & 0 (-)      & 9 (1.22\%)  \\
Dense Veg.  & 10 (0.01\%)  & 11 (0.94\%)  & 12 (1.54\%)   & 0 (-)  \\ 
\hline
No change & & & &  0 (92.95\%) \\
\hline
\end{tabular}
\end{table}
\vspace{-10pt}

The dataset is split using a ratio of 70:15:15 (3653, 784, 784 images) into training, validation, and test sets without grouping by city, ensuring better generalizability across different urban environments.

\subsection{Data Acquisition}
\begin{figure}[h!]
\centering
\includegraphics[width=0.8\linewidth]{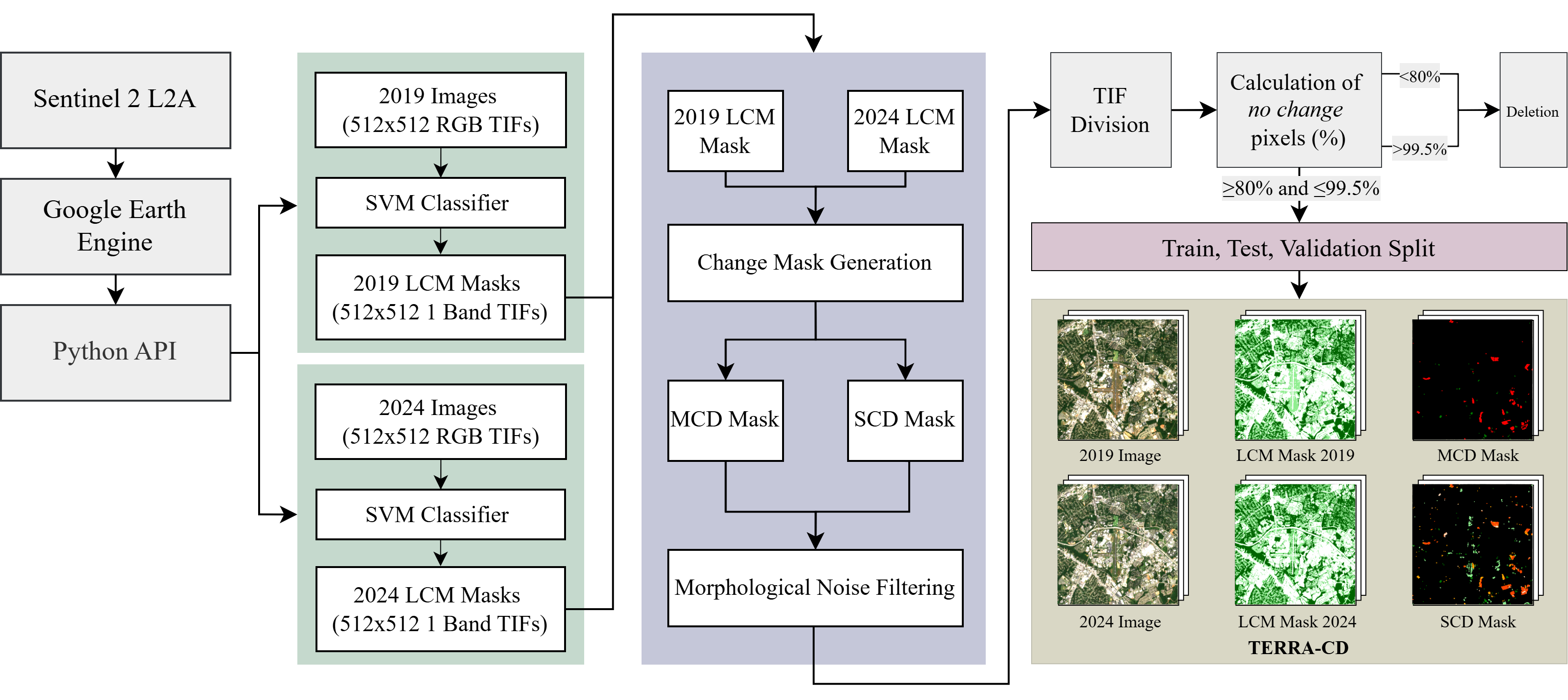}
\caption{Dataset generation process of TERRA-CD}
\label{fig:cd-flowchart}
\end{figure}
\vspace{-10pt}

Figure \ref{fig:cd-flowchart} contains an overview of the dataset generation process of TERRA-CD. We use Google Earth Engine's Python API and computational framework to download imagery for TERRA-CD and for our mask generation process. To ensure the dataset captures correct urban areas, we apply a population threshold of greater than 50,000. Thus, the cities included in TERRA-CD have sufficient urban and vegetation diversity for meaningful analysis. The dataset is based on imagery collected by the Copernicus Sentinel-2 mission \cite{copernicus-s2}. Sentinel-2 provides multispectral imagery with a resolution of 10 meters, ideal for vegetation monitoring and urban analysis. We use the Level-2A (L2A) product, which provides atmospherically corrected Surface Reflectance (SR) imagery ensuring more accurate SR values for reliable change detection.\par

\subsection{LCM Mask Generation}{\label{sec:LCM}}
Normalized Difference Vegetation Index (NDVI) \cite{rouse1974monitoring} and Sentinel Water Mask (SWM) \cite{milczarek2017sentinel} indices are used for land cover classification. These indices provide complementary strengths as NDVI measures vegetation density, distinguishing between sparse tree cover and dense forests, while SWM accurately identifies water bodies. Together, NDVI and SWM are useful for mapping diverse landscapes by clearly separating vegetation, water, and built-up areas.

\begin{figure}[h!]
\centering
\includegraphics[width=0.9\linewidth]{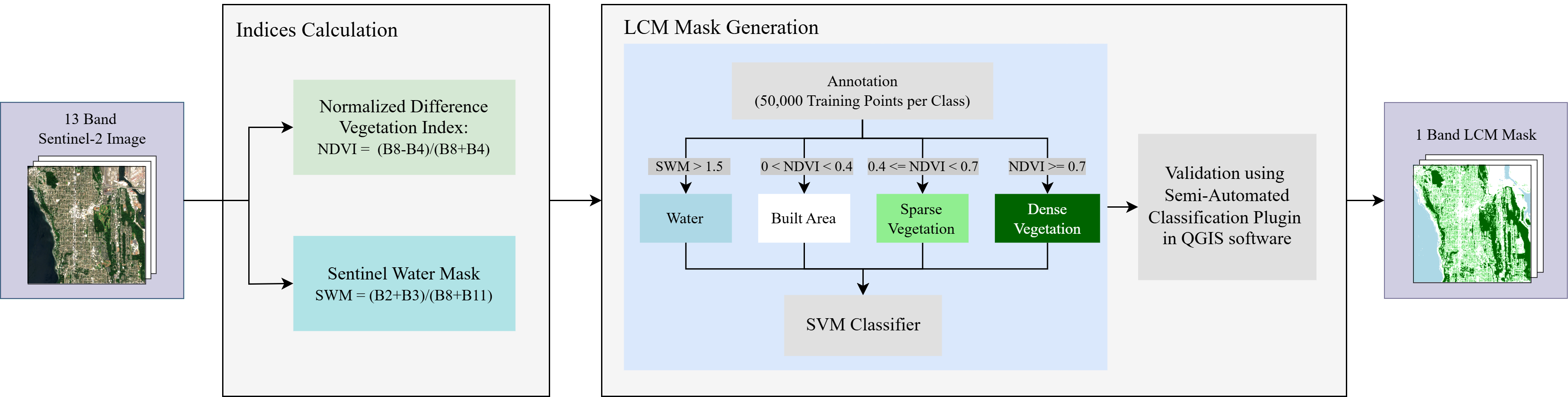}
\caption{Outline of the LCM Mask generation process}
\label{fig:maskgen}
\end{figure}   

The LCM mask generation process is depicted in Figure \ref{fig:maskgen}. We use an automated Support Vector Machine (SVM) classifier for land cover classification of satellite imagery. For each image, we annotate 50,000 points per class as training data using Google Earth Engine. These points are annotated using NDVI thresholds -- NDVI $< 0.4$ for \textit{built area}, between $0.4$ and $0.7$ for \textit{sparse vegetation}, $> 0.7$ for \textit{dense vegetation}, and SWM $> 1.5$ for \textit{water}. These thresholds are determined through iterative refinement informed by the spectral properties of urban and vegetative areas.\par 
The classifier maps every pixel in the image into one of four land cover classes, generating LCM masks. This process is performed independently for 2019 and 2024 images, creating bitemporal mask pairs. We utilize the Spectral Angle Mapping algorithm \cite{liu2013kernel} from Semi-Automatic Classification Plugin (SCP) \cite{congedo2021semi} available in QGIS to manually validate samples from the labeled data and perform ground truthing.

\subsection{Change Mask Generation}\label{sec:maskgen}
In this subsection, we detail the change mask generation process for MCD and SCD tasks. We implement a pixel-based classification method, comparing land cover masks from 2019 and 2024 to observe temporal changes and generate comprehensive change detection masks. For change classification, each transition is systematically evaluated and assigned a class label, as outlined in Table \ref{tab:SCD-class-description}, ensuring a consistent approach to change detection. The classification approach for MCD masks employs a three-class system: \textit{no change, vegetation increase}, and \textit{vegetation decrease}. This method supports analysis of vegetation changes, moving beyond simplistic BCD. For generation of SCD masks, we employ a 13-class system that captures all semantic transitions between \textit{water, built area, sparse } and \textit{dense vegetation}, as detailed in Table \ref{tab:SCD-class-description}. It also presents the distribution of pixels across different land cover transitions in our SCD masks. The percentage of \textit{no change} pixels ($92.95\%$) is considerably less than current datasets (around $99\%$), which is crucial for addressing the class imbalance problem.

\subsection{Preprocessing}{\label{sec:preproc}} 
We implement two techniques to address noise and data quality challenges: morphological noise filtering and histogram-based data cleaning.\par
Using Morphological noise filtering we effectively remove small, isolated pixel clusters (salt-and-pepper noise) while preserving the overall image structure. Equation \ref{eq:morph4} gives the formulation for morphological opening, where $I$ represents the input image, $K$ represents the kernel of size $3 \times 3$, $\circ$ denotes the morphological opening operation, $\ominus$ denotes erosion, $\oplus$ denotes dilation. For a pixel at position $(x,y)$, the erosion and dilation are expressed as $(I \ominus K)(x,y) = \min_{(s,t) \in K} I(x+s,y+t)$ and $(I \oplus K)(x,y) = \max_{(s,t) \in K} I(x-s,y-t)$ respectively.

Therefore, the complete morphological opening operation for noise reduction can be written as shown in Equation \ref{eq:morph4}, where $(x,y)$ represents pixel coordinates in the image, $(s,t)$ and $(u,v)$ represent coordinates in the structuring element $K$, and $K$ is a $3 \times 3$ rectangular structuring element. 

\begin{equation}
(I \circ K)(x,y) = \max_{(s,t) \in K} \min_{(u,v) \in K} I(x-s+u,y-t+v)
\label{eq:morph4}
\end{equation}

The data cleaning process involves calculating the percentage of \textit{no change} pixels in each mask, establishing threshold boundaries between 80\% and 99.5\%, and identifying and removing masks with extreme pixel distributions. Tiles with excessive \textit{no change} pixels are removed as they provide minimal useful information. Similarly, tiles with insufficient \textit{no change} pixels are eliminated as they likely contain noise or registration errors.

The filtered dataset $\Phi$, derived after noise reduction is defined in Equation \ref{eq:hist1}, where $\Omega$ is the set of all masks, $\omega$ represents an individual mask, $M_{\omega}(i,j)$ denotes the pixel value at position $(i,j)$ in mask $\omega$, $I_{(M_{\omega}(i,j)=0)}$ is the indicator function for zero-valued pixels, $\mu(\omega) = m \times n$ represents the total number of pixels in mask $\omega$, and $\lambda_{\text{min}} = 80\%$ and $\lambda_{\text{max}} = 99.5\%$ are the threshold boundaries. 

\begin{equation}
\Phi = { \omega \in \Omega : \lambda_{\text{min}} \leq \frac{\sum_{i=1}^{m} \sum_{j=1}^{n} I_{(M_{\omega}(i,j)=0)}}{\mu(\omega)} \cdot 100 \leq \lambda_{\text{max}} } 
\label{eq:hist1}
\end{equation}

\begin{equation}
\begin{gathered}
    \Delta(\Omega) = { \psi_{\omega} : \omega \in \Omega, \psi_{\omega} = \frac{\sum_{i,j} \beta(M_{\omega}(i,j))}{\mu(\omega)} \cdot 100 }\\ \\
    \beta(x) = \begin{cases} 1 & \text{if } x = 0 \\ 0 & \text{otherwise} \end{cases}
\end{gathered}
\label{eq:hist2}
\end{equation}

The statistical distribution $\Delta$ of zero-pixel percentages across all masks is characterized as shown in Equation \ref{eq:hist2}. Finally, the set of filtered masks $\eta$ is derived using $\eta = { \omega \in \Omega : \psi_{\omega} \notin [\lambda_{\text{min}}, \lambda_{\text{max}}] }$ where $|\eta|$ represents the number of masks exceeding the threshold boundaries. The histogram analysis eliminates masks with minimal detectable changes, excessive noise, and statistically unreliable pixel distributions.\par

\section{Methodology}
\subsection{Training Setup}{\label{sec:train}}
The training is conducted on an NVIDIA GeForce RTX 2080 Ti GPU with CUDA 12.1, using PyTorch 2.1.0. We use a batch size of 32 for our experiments, maintaining a balance between training efficiency and GPU memory constraints. We also implement model checkpointing using validation loss to save the model state at regular intervals throughout the training process.

\subsubsection{Class Weighting Strategy}
To address class imbalance in our dataset, we implement Cross Entropy Loss with class weights calculated using inverse frequency square root \cite{cui2019class}. The Weighted Categorical Cross Entropy Loss $\mathcal{L}_{CE}$ is defined in Equation \ref{eq:ce-loss}, where, $p_{n,k}$ represents the probability for class $k$ at pixel $n$, derived from the model's logit output $z_{n,k}$. The binary indicator $y_{n,k}$ equals 1 when pixel $n$ belongs to class $k$, and 0 otherwise. 
\vspace{-4pt}
\begin{equation}
  \mathcal{L}_{CE} = -\frac{1}{N} \sum_{n=1}^N \sum_{k=1}^K w_k \cdot y_{n,k} \cdot \log(p_{n,k}) \qquad \text{where } p_{n,k} = \frac{e^{z_{n,k}}}{\sum_{j=1}^K e^{z_{n,j}}}
\label{eq:ce-loss}
\end{equation}

For a dataset with $K$ classes, the square balanced weights $w_k$ are calculated using Equation \ref{eq:weights}, where $f_k = \frac{n_k}{N}$ represents the frequency of class $k$, calculated as the ratio of pixels in class $k$ ($n_k$) to the total pixel count ($N$). The normalization term ensures the weights sum to the total number of classes $K$, establishing a balanced distribution across all classes.

\begin{equation}
w_k = \frac{K \sqrt{1/f_k}}{\sum_{i=1}^K \sqrt{1/f_i}} 
\label{eq:weights}
\end{equation}

\subsubsection{Optimization and Learning Rate Scheduling}
For training, we implement the AdamW optimizer \cite{loshchilov2019decoupled} with an initial learning rate of 1e-4 and weight decay of 0.01. A \texttt{ReduceLROnPlateau} scheduler dynamically adjusts the learning rate.

\begin{equation}
\label{eq:lr_schedule}
    \alpha_{new} = \begin{cases} 0.5\alpha_{old} & \text{if } \min(L_{t-p:t}) \geq \min(L_{1:t-p-1}) \\ \alpha_{old} & \text{otherwise} \end{cases}
\end{equation}

The learning rate scheduler updates $\alpha$ after every $p=5$ epochs (patience) if the validation loss hasn't improved, as shown in Equation \ref{eq:lr_schedule}, where $L_{t}$ represents the validation loss at epoch $t$. 

\subsubsection{Evaluation}
To evaluate change detection performance, we employ standard metrics derived from the confusion matrix. In a confusion matrix $\mathbf{C}$, each element $C_{ij}$ refers to how many pixels from true class $i$ are labeled as class $j$ by the model. For the 3-class vegetation MCD task, we report unweighted averages of accuracy, F1-score, mIoU, and kappa coefficient to ensure equal consideration of all change categories. For the SCD evaluation, we employ weighted averages of these metrics to address the class imbalance in land cover transitions. 

\section{Results}
\begin{figure}[h!]
    \centering
    \includegraphics[width=0.8\linewidth]{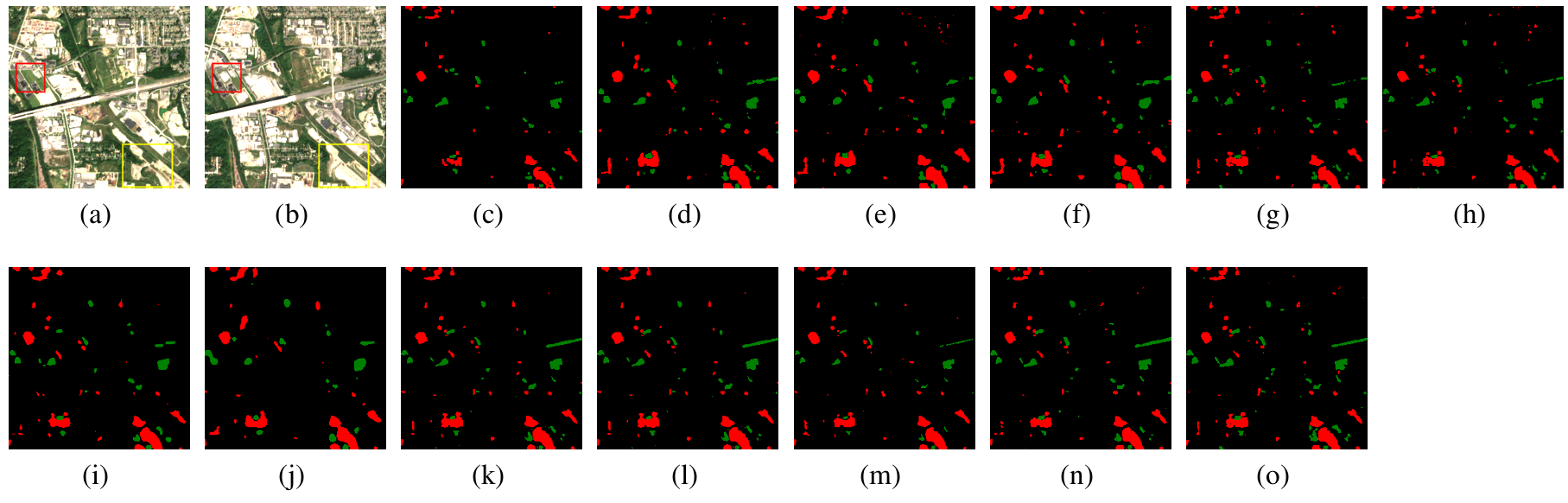}
    \caption{Model Predictions for MCD Masks of Cleveland, Ohio}
    \label{fig:mcd-usa}
\end{figure}

{\label{app:results}}
For the MCD task, we implement multi-class variants of Siamese networks \cite{daudt2018fully}, SNUNet variants \cite{fang2021snunet}, and STANet variants \cite{chen2020spatial}. We also implement PCC approaches using UNet \cite{ronneberger2015unet}, LinkNet \cite{chaurasia2017linknet}, DeepLabv3+ \cite{chen2018encoder}, and PSPNet \cite{zhao2017pyramid}. The performance results of all the models are given in Table \ref{tab:mcd-3class} and qualitative results are presented in Figure \ref{fig:mcd-usa}. \par
\begin{figure}[h!]
    \centering
    \includegraphics[width=0.8\linewidth]{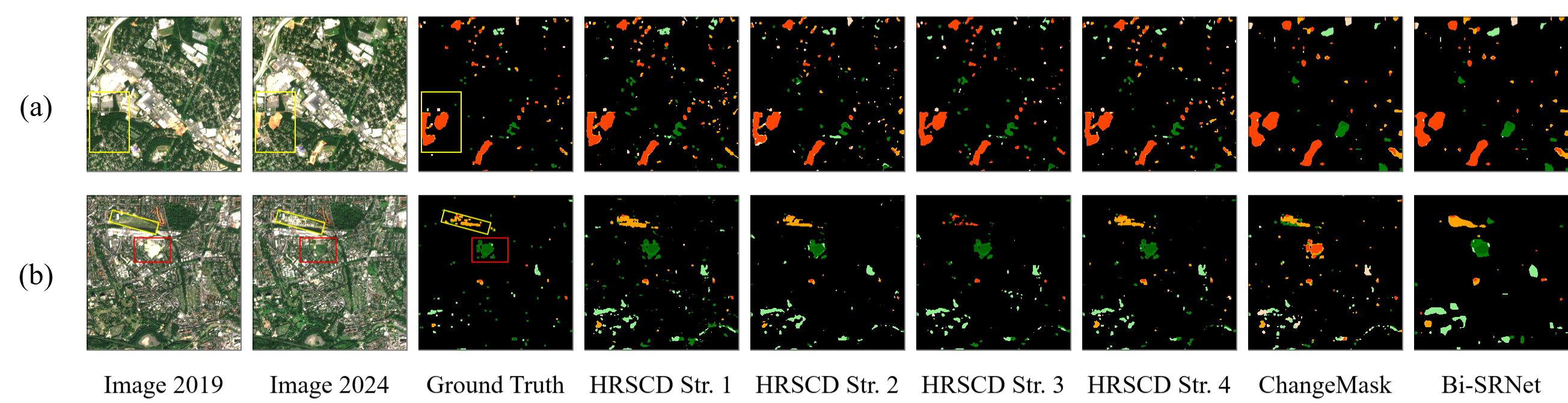}
    \caption{Model predictions for SCD Masks: (\textbf{a}) Charlotte, North Carolina and (\textbf{b}) Dortmund, Germany}
    \label{fig:scd-usa}
\end{figure}

For the SCD task, we implement the four HRSCD strategies proposed by Daudt et al. \cite{daudt2019multitask}, Bi-SRNet \cite{ding2022bitemporal} and Changemask \cite{zheng2022changemask}. The SCD Change Detection results are given in Table \ref{tab:scd-13class}. Additionally, the qualitative results are shown in Figure \ref{fig:scd-usa}.\par
Our benchmarking results establish TERRA-CD as a dataset useful for change detection research. The performance metrics across all evaluated models fall within comparable ranges to those reported on other prominent change detection datasets such as HRSCD \cite{daudt2019multitask}, SECOND \cite{yang2020semantic}, and Hi-UCD \cite{tian2022large}. 

\begin{table}[h!]
\small
\centering
\setlength\tabcolsep{1pt}
\caption{3-class MCD Results}
\label{tab:mcd-3class}
\begin{tabularx}{0.8\linewidth}{l @{\extracolsep{\fill}} c c c c}
\hline
\textbf{Model} & \textbf{OA(\%)} & \textbf{F1(\%)} & \textbf{Kappa(\%)} & \textbf{mIOU(\%)} \\
\hline
Multi STANet-Base  & 97.21 & 75.69 & 63.27 & 63.94 \\
Multi STANet-BAM   & 97.26 & 75.86 & 63.66 & 64.13 \\
Multi STANet-PAM   & 97.17 & 75.12 & 62.36 & 63.32 \\
\hline
UNet               & 97.91 & 78.33 & 69.20 & 67.56 \\
LinkNet            & 98.06 & 79.37 & 70.64 & 68.73 \\
DeepLabV3+         & 97.37 & 74.45 & 63.35 & 63.15 \\
PSPNet             & 97.10 & 72.08 & 60.14 & 60.64 \\
\hline
Multi FC-Siam-conc & 97.48 & 77.33 & 65.89 & 65.79 \\
Multi FC-Siam-diff & 97.75 & 78.12 & 67.62 & 66.74 \\
Multi FC-EF        & 97.79 & 77.67 & 66.56 & 66.23 \\
Multi SNUNet-conc  & 97.76 & 78.88 & 68.63 & 67.64 \\
Multi SNUNet-ECAM\;  & 97.59 & 77.97 & 67.14 & 66.54 \\
\hline
\end{tabularx}
\end{table}
\vspace{-20pt}

\vspace{-10pt}

\begin{table}[h!]
\small
\centering
\setlength\tabcolsep{1pt}
\caption{13-class SCD Change Detection Results}
\label{tab:scd-13class}
\begin{tabularx}{0.8\linewidth}{@{\extracolsep{\fill}} lcccc}
\hline
\textbf{Model} & \textbf{OA(\%)} & \textbf{F1(\%)} & \textbf{mIoU(\%)} & \textbf{Kappa(\%)} \\
\hline
HRSCD Str. 1 & 93.06 & 94.11 & 70.32 & 58.79 \\
HRSCD Str. 2 & 93.92 & 94.28 & 68.84 & 59.98 \\
HRSCD Str. 3 & 93.50 & 93.44 & 68.31 & 55.53 \\
HRSCD Str. 4 & 93.94 & 94.50 & 71.83 & 61.51 \\
Bi-SRNet & 90.38 & 91.86 & 65.89 & 45.71 \\
ChangeMask & 92.56 & 93.24 & 65.93 & 52.67 \\
\hline
\end{tabularx}
\end{table}

\section{Conclusion}
In this paper, we introduce TERRA-CD, a comprehensive benchmark dataset for vegetation change detection spanning diverse urban environments across the USA and Europe. With 5,221 high-quality image pairs and a multi-granular classification approach supporting both 3-class MCD and 13-class SCD, TERRA-CD addresses critical gaps in existing resources. Our benchmarking results validate TERRA-CD as a versatile change detection benchmark dataset, useful for improving current models and developing novel architectures for MCD and SCD. \par
Future work will focus on manual validation of all LCM masks and integration with multimodal data sources such as SAR and LiDAR. As climate change intensifies and urbanization continues, we believe TERRA-CD will accelerate progress in satellite-based environmental monitoring, enabling more accurate tracking of urban development and vegetation changes to support decision-making for sustainable development.



\begin{thebibliography}{6}
\bibitem{chen2020spatial}Chen, H. \& Shi, Z. A spatial-temporal attention-based method and a new dataset for remote sensing image change detection. {\em Remote Sensing}. \textbf{12} pp. 1662 (2020)
\bibitem{adriano2021learning}Adriano, B., Yokoya, N., Xia, J., Miura, H., Liu, W. \& Matsuoka, M. Learning from multimodal and multitemporal earth observation data for building damage mapping. {\em ISPRS Journal Of Photogrammetry And Remote Sensing}. \textbf{175} pp. 132-143 (2021)
\bibitem{daudt2019multitask}Daudt, R., Le Saux, B., Boulch, A. \& Gousseau, Y. Multitask learning for large-scale semantic change detection. {\em Computer Vision And Image Understanding}. \textbf{187} pp. 102783 (2019)
\bibitem{yang2020semantic}Yang, K., Xia, G., Liu, Z., Du, B., Yang, W., Pelillo, M. \& Zhang, L. Asymmetric siamese networks for semantic change detection in aerial images. {\em IEEE Transactions On Geoscience And Remote Sensing}. \textbf{60} pp. 1-18 (2021)
\bibitem{zhu2024review}Zhu, Q., Guo, X., Li, Z. \& Li, D. A review of multi-class change detection for satellite remote sensing imagery. {\em Geo-spatial Information Science}. \textbf{27}, 1-15 (2024)
\bibitem{lyu2016learning}Lyu, H., Lu, H. \& Mou, L. Learning a Transferable Change Rule from a Recurrent Neural Network for Land Cover Change Detection. {\em Remote Sensing}. \textbf{8} pp. 506 (2016)
\bibitem{du2019unsupervised}Du, B., Ru, L., Wu, C. \& Zhang, L. Unsupervised Deep Slow Feature Analysis for Change Detection in Multi-Temporal Remote Sensing Images. {\em IEEE Transactions On Geoscience And Remote Sensing}. \textbf{57} pp. 9976-9992 (2019)
\bibitem{wu2016scene}Wu, C., Zhang, L. \& Zhang, L. A Scene Change Detection Framework for Multi-Temporal Very High Resolution Remote Sensing Images. {\em Signal Processing}. \textbf{124} pp. 184-197 (2016)
\bibitem{song2018change}Song, A., Choi, J., Han, Y. \& Kim, Y. Change Detection in Hyperspectral Images Using Recurrent 3D Fully Convolutional Networks. {\em Remote Sensing}. \textbf{10} pp. 1827 (2018)
\bibitem{liu2019review}Liu, S., Marinelli, D., Bruzzone, L. \& Bovolo, F. A Review of Change Detection in Multitemporal Hyperspectral Images: Current Techniques, Applications, and Challenges. {\em IEEE Geoscience And Remote Sensing Magazine}. \textbf{7} pp. 140-158 (2019)
\bibitem{tian2022large}Tian, S., Zhong, Y., Zheng, Z., Ma, A., Tan, X. \& Zhang, L. Large-scale deep learning based binary and semantic change detection in ultra high resolution remote sensing imagery: From benchmark datasets to urban application. {\em ISPRS Journal Of Photogrammetry And Remote Sensing}. \textbf{193} pp. 164-186 (2022)
\bibitem{hussain2013change}Hussain, M., Chen, D., Cheng, A., Wei, H. \& Stanley, D. Change detection from remotely sensed images: From pixel-based to object-based approaches. {\em ISPRS Journal Of Photogrammetry And Remote Sensing}. \textbf{80} pp. 91-106 (2013)
\bibitem{copernicus-s2}ESA Copernicus Sentinel-2 Mission Documentation. Available online: \url{https://documentation.dataspace.copernicus.eu/Data/SentinelMissions/Sentinel2.html/}, Accessed 12 Dec 2025
\bibitem{rouse1974monitoring}Rouse, J., Haas, R., Deering, D., Schell, J. \& Harlan, J. Monitoring the Vernal Advancement and Retrogradation (Green Wave Effect) of Natural Vegetation.  (1974), Available online: \url{https://ntrs.nasa.gov/citations/19750020419}, Accessed 14 Dec 2025
\bibitem{milczarek2017sentinel}Milczarek, M., Robak, A. \& Gadawska, A. Sentinel Water Mask (SWM) - New index for water detection on Sentinel-2 Images. {\em Proceedings Of The 7th Advanced Training Course On Land Remote Sensing}. (2017)
\bibitem{liu2013kernel}Liu, X. \& Yang, C. A Kernel Spectral Angle Mapper Algorithm for Remote Sensing Image Classification. {\em Proceedings Of The 2013 6th International Congress On Image And Signal Processing (CISP)}. \textbf{2} pp. 814-818 (2013)
\bibitem{congedo2021semi}Congedo, L. Semi-Automatic Classification Plugin: A Python Tool for the Download and Processing of Remote Sensing Images in QGIS. {\em Journal Of Open Source Software}. \textbf{6}, 3172 (2021)
\bibitem{cui2019class}Cui, Y., Jian, M., Lin, T., Song, Y. \& Belongie, S. Class-Balanced Loss Based on Effective Number of Samples. {\em IEEE/CVF Conference On Computer Vision And Pattern Recognition (CVPR)}. pp. 9268-9277 (2019)
\bibitem{loshchilov2019decoupled}Loshchilov, I. \& Hutter, F. Decoupled Weight Decay Regularization. {\em International Conference On Learning Representations (ICLR)}. (2019)
\bibitem{daudt2018fully}Daudt, R., Le Saux, B. \& Boulch, A. Fully convolutional siamese networks for change detection. {\em IEEE International Conference On Image Processing (ICIP)}. pp. 4063-4067 (2018)
\bibitem{fang2021snunet}Fang, S., Li, K., Shao, J. \& Li, Z. SNUNet-CD: A densely connected Siamese network for change detection of VHR images. {\em IEEE Geoscience And Remote Sensing Letters}. \textbf{19} pp. 1-5 (2021)
\bibitem{ronneberger2015unet}Ronneberger, O., Fischer, P. \& Brox, T. U-Net: Convolutional networks for biomedical image segmentation. {\em Medical Image Computing And Computer-Assisted Intervention (MICCAI)}. pp. 234-241 (2015)
\bibitem{chaurasia2017linknet}Chaurasia, A. \& Culurciello, E. LinkNet: Exploiting encoder representations for efficient semantic segmentation. {\em IEEE Visual Communications And Image Processing (VCIP)}. pp. 1-4 (2017)
\bibitem{chen2018encoder}Chen, L., Zhu, Y., Papandreou, G., Schroff, F. \& Adam, H. Encoder-decoder with atrous separable convolution for semantic image segmentation. {\em European Conference On Computer Vision (ECCV)}. pp. 801-818 (2018)
\bibitem{zhao2017pyramid}Zhao, H., Shi, J., Qi, X., Wang, X. \& Jia, J. Pyramid Scene Parsing Network. {\em IEEE Conference On Computer Vision And Pattern Recognition (CVPR)}. pp. 2881-2890 (2017)
\bibitem{ding2022bitemporal}Ding, L., Guo, H., Liu, S., Mou, L., Zhang, J. \& Bruzzone, L. Bi-temporal semantic reasoning for the semantic change detection in HR remote sensing images. {\em IEEE Transactions On Geoscience And Remote Sensing}. \textbf{60} pp. 1-14 (2022)
\bibitem{zheng2022changemask}Zheng, Z., Zhong, Y., Tian, S., Ma, A. \& Zhang, L. ChangeMask: Deep multi-task encoder-transformer-decoder architecture for semantic change detection. {\em ISPRS Journal Of Photogrammetry And Remote Sensing}. \textbf{183} pp. 228-239 (2022)

\end{thebibliography}
\end{document}